\documentclass[authoryear,A4]{elsarticle}

\usepackage{vmargin}

\setmargins
{2.5cm}				
{1.5cm}				
{16cm}				
{25cm}				
{10pt}				
{1cm}				
{0pt}				
{1cm}				

\usepackage[utf8]{inputenc}
\usepackage{amssymb}
\usepackage{lineno}
\usepackage{url}
\usepackage{hyperref}
\hypersetup{unicode=true,
           colorlinks=true,
           linkcolor=blue,
           citecolor=blue,
           filecolor=red,
           urlcolor=blue,
           breaklinks=true
           }

\journal{arXiv}

\begin{document}

\begin{frontmatter}

\title{Image Classification of Grapevine Buds using Scale-Invariant Features Transform, Bag of Features and Support Vector Machines}

\author[utn,uncuyo]{Diego Sebastián Pérez\corref{cor1}}
\ead{sebastian.perez@frm.utn.edu.ar}

\author[conicet]{Facundo Bromberg}
\ead{fbromberg@frm.utn.edu.ar}

\author[utn]{Carlos Ariel Diaz}
\ead{carlos.diaz@frm.utn.edu.ar}

\address[utn]{Universidad Tecnológica Nacional, Facultad Regional Mendoza, Laboratorio de Inteligencia Artificial DHARMa, Dpto. de Sistemas de la Información. Rodríguez 273, CP 5500, Mendoza, Argentina.}

\address[uncuyo]{Universidad Nacional de Cuyo, Instituto universitario para las Tecnologías de la Información y las Comunicaciones, CONICET. Padre Jorge Contreras 1300, CP 5500, Mendoza, Argentina.}

\address[conicet]{Universidad Tecnológica Nacional, Facultad Regional Mendoza, CONICET, Laboratorio de Inteligencia Artificial DHARMa, Dpto. de Sistemas de la Información. Rodríguez 273, CP 5500, Mendoza, Argentina.}

\cortext[cor1]{Corresponding author}

\begin{abstract}
In viticulture, there are several applications where bud detection in vineyard images is a necessary task, susceptible of being automated through the use of computer vision methods. A common and effective family of visual detection algorithms are the \emph{scanning-window} type, that slide a (usually) fixed size window along the original image, classifying each resulting windowed-patch as containing or not containing the target object. The simplicity of these algorithms finds its most challenging aspect in the classification stage. Interested in grapevine buds detection in natural field conditions, this paper presents a classification method for images of grapevine buds ranging 100 to 1600 pixels in diameter, captured in outdoor, under natural field conditions, in winter (i.e., no grape bunches, very few leaves, and dormant buds), without artificial background, and with minimum equipment requirements. The proposed method uses well-known computer vision technologies: \emph{Scale-Invariant Feature Transform} for calculating low-level features, \emph{Bag of Features} for building an image descriptor, and \emph{Support Vector Machines} for training a classifier.
When evaluated over images containing buds of at least 100 pixels in diameter, the approach achieves a recall higher than $0.9$ and a precision of $0.86$ over all windowed-patches covering the whole bud and down to 60\% of it, and scaled up to window patches containing a proportion of 20\%-80\% of bud versus background pixels. This robustness on the position and size of the window demonstrates its viability for use as the classification stage in a scanning-window detection algorithms.
\end{abstract}


\end{frontmatter}

\section{Introduction}
\label{sec:intro}

There is a vast number of \emph{computer vision algorithms} used in a variety of applications for different areas of human activity \citep{szeliski2010computer,chen2010handbook,ikeuchi2014computer,guo2015deep}, among which we can find applications for agricultural activity \citep{mccarthy2010applied,chen2002machine,gomes2012applications,zhang2014application,vibhute2012applications}.
These algorithms are increasingly used in agricultural industry for faster, more economical, and more objective inspection, measurement, and evaluation tasks. In recent decades, advances in hardware and software have motivated a lot of work aimed at developing computer vision systems for agriculture problems, showing that this technology has an enormous potential for automating the guidance and control of agricultural processes. A very active discipline in \emph{agricultural sciences} that uses computer vision technology is \emph{precision viticulture} \citep{whalley2013applications}. Computer vision algorithms are commonly used for acquiring a diverse range of information from the vineyards, such as grapes and bunches detection \citep{nuske2011yield}, estimation of grape size and weight \citep{tardaguila2012automatic}, bud detection \citep{xu2014detection}, leaf area and yield estimation \citep{diago2012grapevine}, plants phenotyping \citep{herzog2015initial}, autonomous selective sprayer \citep{berenstein2010grape}, and more \citep{tardaguila2012applications}. Among all these applications, classification and detection of grapevine buds in natural environment are among the most challenging for computer vision algorithms, while they have a great impact as potential solutions to many problems in viticulture.

In this work, we propose a classification method for images of grapevine buds ranging 100 to 1600 pixels in diameter, taken under natural field conditions in winter (i.e., no grape bunches, very few leaves, and dormant buds), based on \emph{visual feature extraction} and \emph{image classification} techniques, that satisfies the requirements for detecting grapevine buds using \emph{scanning-window object detection} algorithms. Object detection algorithms are designed to detect all object instances of a known class in an image. Usually, only a small number of the object instances are present in the image, but there are a large number of possible locations and scales on which they may appear \citep{ikeuchi2014computer}. Scanning window detection algorithms are a popular family of detection algorithms chosen in practice when it is required a recognition system with high performance \cite{divvala2009empirical}. In scanning-window detection algorithms \citep{divvala2009empirical,wang2009hog}, each image is densely scanned from one end of the image (e.g. upper left corner) to the other end (e.g. bottom right corner) with rectangular \emph{sliding windows} on different scales, extracting sub-images or \emph{patches} of the original image. For each patch, different visual features are extracted (e.g. edges, regions, textures, gradients, and more) and fed to a classifier that has been previously trained using patches labeled in their respective classes. Typical examples of classifiers are support vector machines, decision trees, artificial neural network, random forest, statistical classification, and more. The classifier discriminates the new patches containing the desired object as positives, and the rest as negative cases. Finally, the detection of a particular object occurs when some number of patches in some region of the image were classified as containing the object.
So we see that in a scanning-window detection scheme, the role of the classification algorithm is essential, relegating the rest of the implementation details, such as size and shape of window, displacement and scale, to decisions of design particular to the detection problem to be solved.

A mature buds detection technology presents important application opportunities, such as \emph{grapevine pruning}, \emph{grapevine plant phenotyping}, and \emph{3D reconstruction of the plant structure and components}, among others. We discuss these applications in some detail to help us better assess the potential impact of our contribution. 

Grapevine pruning is a key task in managing vineyards that seeks the balance between productivity and quality of the plant fruits \citep{keller2015science}. Manual operation is the main method of pruning, requiring hard physical labor during several hours. Different \emph{dormant pruning} strategies such as \emph{spur pruning} and \emph{cane pruning} \citep{wolf1995mid}, adjust the levels of pruning of the plant branches, based on the counting of (dormant) buds that needs to be retained after pruning. Generally, this count does not include basal buds, i.e. non-fruitful buds that develop at the base of a shoot, closest to the cordon. We can see that pruning requires skilled operators for identifying the buds that conform to the specifications set by the method, as well as deciding the recommended position for the cut. The need for skilled operators makes pruning one of the most expensive tasks in the management of vineyards \citep{billikopf1992pay}. It is possible to reduce pruning costs between 56\% and 80\% by fully mechanizing this task \citep{bates2009mechanical}, or by implementing pruning expert systems to assist less-skilled operators \citep{corbett2012expert}. Such a system should be able to count and identify different buds, which involves classification, detection, 3D reconstruction, and other complex operations. Depending on the pruning strategy, the system requires robust detection of buds, meaning high values of recall (or bud detection rate). Detection errors by low values of recall implies overlooked buds, leading to misconceptions about where the cut should be made. To date, we found in the literature some preliminary studies that put their attention on this problem \citep{xu2014detection,corbett2012expert,gao2006image}, but these are approximate solutions, i.e. under controlled conditions or in environments simulated computationally. There is also the \emph{training pruning}, performed at different developmental stages of the plant. While all these types of pruning are different from the agricultural point of view, mechanizing them requires the same automatic buds detection technologies.

Another potential application for buds detection is the phenotyping of plants, a tool increasingly used in viticulture. Through phenotyping, it is possible to carry out a quantitative assessment of the plants characteristics, to obtain its anatomical and physiological description \citep{berger2012high}. Generally, phenotyping is carried out by direct visual inspection of the plant, so it is strongly limited by time, costs, and operator subjectivity. Currently, there are high performance, non-destructive phenotyping platforms based on processing and image analysis techniques, that although present important advances in recent years with \emph{real life applications} under field conditions \citep{kicherer2015automated,klodt2015field}, are limited to obtain an initial characterization for some plant features \citep{walter2015plant,hartmann2011htpheno}. For grapevines, the buds development stage during sprouting is one of their phenotypic traits. Therefore, high performance phenotyping becomes closer to reality when autonomous classification and detection of buds is mature enough to distinguish different levels of development. According to the results obtained by \cite{herzog2015initial}, a recall of 96\% for buds detection in development stage BBCH\footnote{In biology, BBCH scale (Biologische Bundesanstalt, Bundessortenamt und CHemische Industrie) for vineyards describes the phenological development of the grapevine.} 10 is enough to achieve the requirements of grapevines phenotyping during sprouting stage, whose coding in BBCH scale covers descriptors from 00 to 10 \citep{lorenz1995growth}. In our work, we did not discriminate buds over their development stages, but we can recognize, through manual inspection, that the images we used range from 00 to 05 in the BBCH scale \citep{andreini2015study}.

A third possible application for buds detection is the 3D reconstruction of the plant structure: the shape, location, and orientation of its components such as buds themselves, as well as other plant components such grape bunches, leaves, trunk and branches, tendrils, leaf axils. Detecting the 3D position of buds is useful in 3D reconstruction as this position is in direct correlation with the leaf axil 3D position, as in all but few rare cases, buds are located exactly at the leaf axil; and also buds in winter appear at the location where leaves, grape bunches, or tendrils were located in the previous summer. 
Locating the components of the plant may be of direct interest to agronomists. For instance, in grapevines, as in most crops, variations in the interception and distribution of light over the canopy and buds structure, affects productivity, yield, and quality \citep{smart1985principles}.
Also, detecting the 3D location of components of actual plants may be useful to improve simulations of the physiological responses and physical processes mentioned above. In recent years, different simulation methods have emerged for producing approximate 3D plant structure \citep{iandolino2013simulating,louarn2008three}. To acquire more realistic 3D models, simulations may be combined with information obtained from 2D or 3D measurements over the actual plant structure \citep{sievanen2014functional}. These measurements may be obtained autonomously through computer vision algorithms: \emph{object segmentation, classification, and detection} for 2D measurements \citep{kicherer2014image,diago2012grapevine}; and \emph{structure from motion} or \emph{multi-view 3D reconstruction} for 3D measurements \citep{dey2012classification,quan2010image}.
Clearly, the more accurate are the localization of the plant components obtained by classification and detection algorithms, the more accurate is the plant model obtained from the 3D reconstruction. However, misclassification errors that locate a bud where there is none (measured by the detection precision), or that omit a bud where there is one (measured by the detection recall), may increase the errors in the 3D plant model reconstruction; resulting on one hand in a direct transfer of these errors to the plant components directly correlated with the buds such as grape bunches, leaves, tendrils, and leaves axil; but may also have an important impact in the 3D model reconstruction by providing false restrictions.

In this study we present a binary classification algorithm for patches of grapevine buds which satisfies the requirements of a scanning-window detection algorithm, for images of vineyards captured in their natural environment. This approach was build using known computer vision algorithms, such as \emph{Scale-Invariant Feature Transform} \citep{lowe2004distinctive}, \emph{Bag of Features} \citep{csurka2004visual} and \emph{Support Vector Machines} \citep{vapnik1998statistical}. Our approach assumes patches contain at most one bud, and is robust to patches containing at least 60\% of bud pixels, while the bud pixels in the patch are between 20\% and 80\% compared to \emph{non-bud} pixels (i.e. parts of natural environment in vineyard scenes that are not bud). As shown in section \ref{sec:results}, the results achieved by our classification algorithm are robust enough for their use within a bud detection algorithm based on scanning-window. Section \ref{sec:matmet} provides further details on implementation of the proposed approach.

\subsection{Related work}

We discuss in this section some related work in the literature that presents alternative solutions to the same problem, as well as works that presents solutions to similar problems.

To the best of our efforts, we could not find in the literature a single solution to the problem of grapevine buds classification. Instead, we found two state of the art works that solves the problem of grapevine buds detection. \cite{xu2014detection} presents a buds detection algorithm on grapevines, whose main motivation is to set the basis for an autonomous system of grapevine pruning in winter. The buds detection is performed from RGB images (its resolution is not informed), captured with an industrial CCD\footnote{CCD (Charge-Coupled Device) sensors are usually more expensive than CMOS (Complementary Metal Oxide Semiconductor) sensors, which are more common in compact digital cameras.} camera, in indoor, with an artificial white background, and controlled lighting. To discriminate the background and plant pixels, they applied a threshold-based filter on the B (blue) component of the RGB images, resulting in a binary image, which is then subjected to a \emph{thinning process} for obtaining a \emph{wired skeleton} of the plant. Under the assumption that the morphological features of the buds are similar to corners, they applied Harris algorithm \citep{harris1988combined} to the skeleton image for the detection of corners, and discarded the low value corners by some proposed ad-hoc thresholding (details omitted here). This method achieves a recall (a.k.a., true positive rate) of 70.2\%. While the authors explain possible improvements to increase the detection rate, the most important limitation of this approach is the need for imaging under controlled laboratory conditions, i.e. in indoor, with artificial background, and controlled lighting. A second work for buds detection is presented by \cite{herzog2015initial} that introduces three approaches for detecting buds under advanced development (BBCH 10) carried out in experimental vineyards: (a) semi-automatic approach, which requires user interaction to validate the results quality of detection based on RGB images of $3456\times2304$ pixel resolution, with an artificial black background, producing a recall of 94\%; (b) fully autonomous method, that does not require user interaction, also based on RGB images of $3456\times2304$ pixels resolution, with artificial black background, producing 65\% recall; and (c) automated method, that requires the calculation of depth maps (i.e. 2D images containing the relative distances of the surfaces in the scene for the viewpoint), without artificial background, which achieves a recall of 63\%. The authors argue that the recall of the approach (a) is good enough to meet the requirements of plant phenotyping. However, as the authors note, the results of (a) are explained by morphology and colors of buds when photographed. At the beginning of BBCH 10 buds are already starting to sprout, and these sprouts are mostly small, growing near the branch non-uniformly and have similar colors (yellow, light green, beige or brown) to other plant components. Thus, the images used in (a) are captured a few days after the first identification of BBCH 10, when the buds are visibly green and their average size is about 2 cm, resulting in a morphology more distinguishable from other plant components. The implementation details of image processing and stereo vision algorithms used in this work are regrettably not provided.

These works introduce advances in specific applications for buds detection, however, suffer some of the following limitations non existent in our approach: (i) use artificial background in outdoor; (ii) controlled lighting in indoor; (iii) need for user interaction; (iv) specialized stereo vision equipment; or (v) detecting buds in late-stage development, which is expected to be easier to classify because of their greater visual differentiation with the environment. These limitations and the recall reported in these works represent a barrier for advancement in solutions for the applications mentioned in the introduction. 
The literature presents also works based on computer vision algorithms for the detection of grapevine components other than their buds. There are many approaches to perform detection of grapes and bunches of grapevines from RGB images, with applications ranging from autonomous spraying to production estimate.
\cite{nuske2011yield} presents an algorithm based on shape and texture features to detect green grapes, without occlusion, maintaining the natural background, achieving precision of 98\% and recall of 64\%.
\cite{berenstein2010grape} introduces a detector of grapes and grape bunches, based on edge detection and thresholding algorithms achieving a recall for grapes and bunches close to 90\%.
\cite{reis2012automatic} designed a bunches detector for red and white grapes using very simple image processing techniques, such as count of pixels, thresholding and color mapping, with a recall of 91\% for white grapes and 97\% for red grapes.
Another novel application of detection algorithms in viticulture is the assessment of the number of flowers per inflorescence, being one of the key information sources to estimate yield grapevine.
\cite{diago2014assessment} developed a simple, cheap, fast, accurate, and robust approach based on image processing algorithms to autonomously estimate the number of flowers by inflorescence. This approach is applied to RGB images captured in field conditions achieving a precision of 93\% and recall of 74\%.
While there are solutions at different levels for detecting parts of the grapevine, buds classification and detection remains an open problem that is currently under study by the community of precision viticulture and computer vision. On the other hand, we consider that the works discussed here constitute a solid base of evidence on the importance of object detection and image classification algorithms in agricultural sciences and precision viticulture.

\section{Materials and Methods}
\label{sec:matmet}

We present in this section the workflow used to build the classification model, including an explanation of the creation of the training sets, i.e., the images captured and the training patches generated, as well as the implementation details of the classification algorithm.

Our approach aims to solve one of the most complex stages of scanning-window detection algorithms for grapevines buds detection: classification of rectangular image patches as either \emph{bud} patches (containing some part of a bud), or \emph{non-bud} patches (containing not a single bud pixel). We propose a supervised-learning approach for learning the classification model, using as input patches labeled as \emph{bud} or \emph{no-bud}, obtained by a process of scanning-window, over images of representative vineyards taken under natural field conditions, exemplified in Fig.\ref{fig:image-bud}. Our method can be separated into three main parts: generation of the labeled corpus of window patches, calculation of the visual descriptor of each of these patches, and learning the classification model. To generate the labeled corpus, we first captured the images in the field using compact digital cameras. Then we generated a corpus of window patches according to the requirements of scanning-window detection algorithms, and we finished by labeling these patches manually as bud or non-bud. To calculate the visual descriptor of each patch, the algorithm extracts a set of low-level visual features, and builds a representative descriptor of the patch, resulting in a new corpus of labeled descriptors, with one element in the corpus per patch. Finally, the learning of the bud/no-bud classification model consisted in training a supervised binary classifier taking as input a subset of this labeled corpus, called hereon the \emph{train set}. The remaining labeled patches, called hereon the \emph{test set}, are used later on in the experimental section for evaluation of the classification model. We explain in the following sub-section each of these stages in more detail. 

\subsection{Corpus of labeled Patches}
\label{subsec:datasets}

\begin{figure}
    \centering
    \includegraphics[width=12cm]{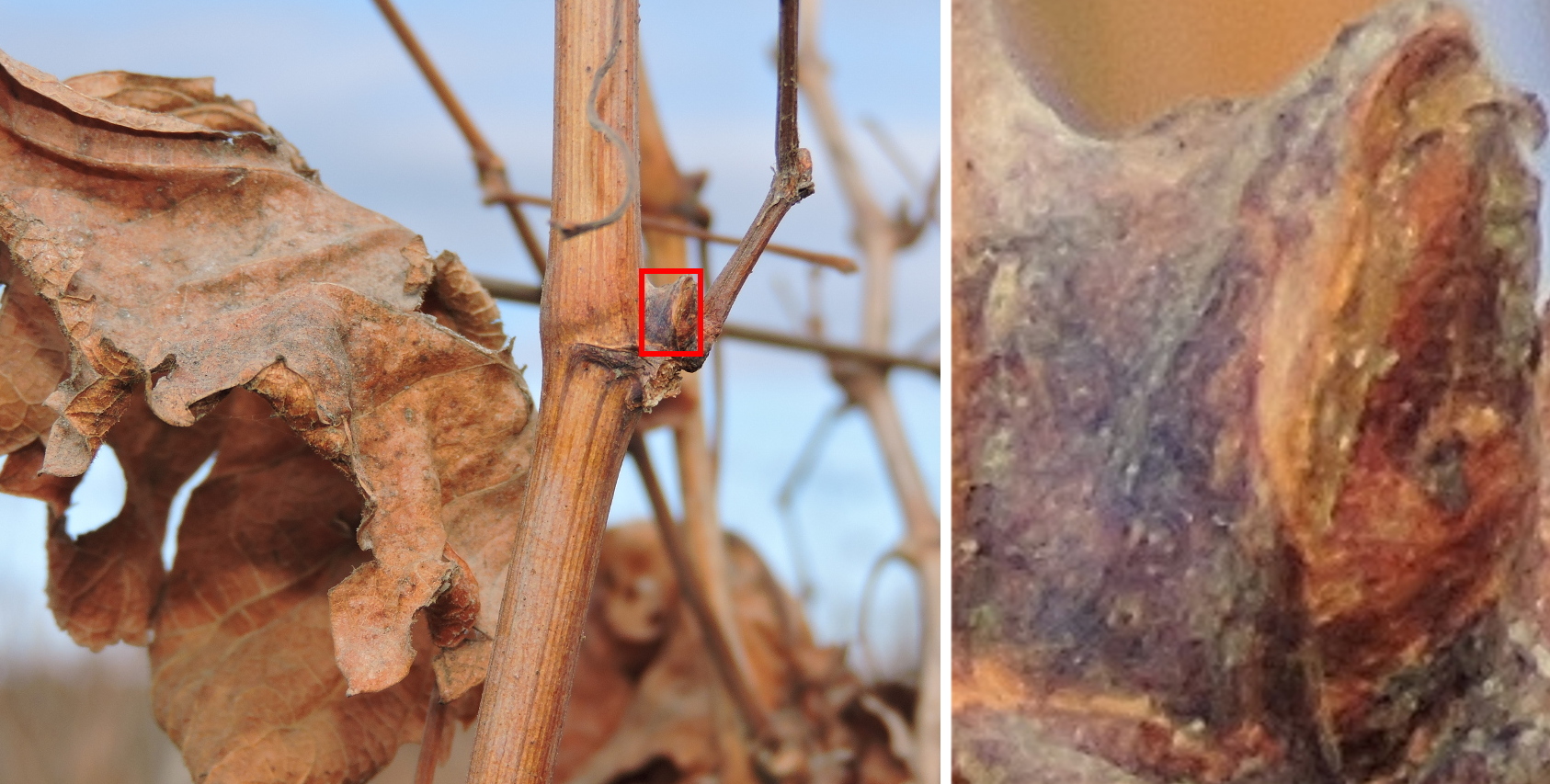}
    \caption{Left: high-resolution image captured in the field. Right: bud patch extracted from the image on the left (marked by the bounding box).}
    \label{fig:image-bud}
\end{figure}

The corpus of labeled patches used in our study was generated with the aim of building a classifier that learns the concept of bud versus non-bud. In this work, the non-bud class is everything else that is not a bud, e.g. stems, leaves, trunks, wires, soil, sky, and more. In this scenario, our approach aims to be used by a scanning-window detection algorithm to detect buds whose diameter ranges from 100 to 1600 pixels within high-resolution images captured with compact digital cameras. The high-resolution images were captured in vineyards of the Department of Agricultural Sciences, Universidad Nacional de Cuyo (National University of Cuyo), and in vineyards of the Instituto Nacional de Tecnología Agropecuaria (National Agricultural Technology Institute), Lujan de Cuyo branch, Mendoza, Argentina. These images were taken with different devices: (1) compact camera Nikon Coolpix P530; (2) compact reflex camera Samsung NX1000; and (3) mobile camera Samsung SM-G920I. All images were captured in JPEG format with a resolution ranging from 12Mpx to 20Mpx, depending on the device used. The images were captured under normal field conditions, without altering the scene with artificial elements, and maintaining the conditions of natural lighting. Only for some cases the flash camera was activated to produce a set of images with greater variability. In total, 760 images were captured containing one or more buds per image, together with other elements of vineyards and its natural environment, such as stems, leaves, trunks, wires, soil, other plants, sky, and more (see Fig.\ref{fig:image-bud}). The images were captured between 10am and 4pm, in different campaigns between July 10 and September 24, 2015 (second half of Argentina's winter), when the plant leaves are dry or have been fallen, but before they begun to sprout again.

\begin{figure}
    \centering
    \includegraphics[width=12cm]{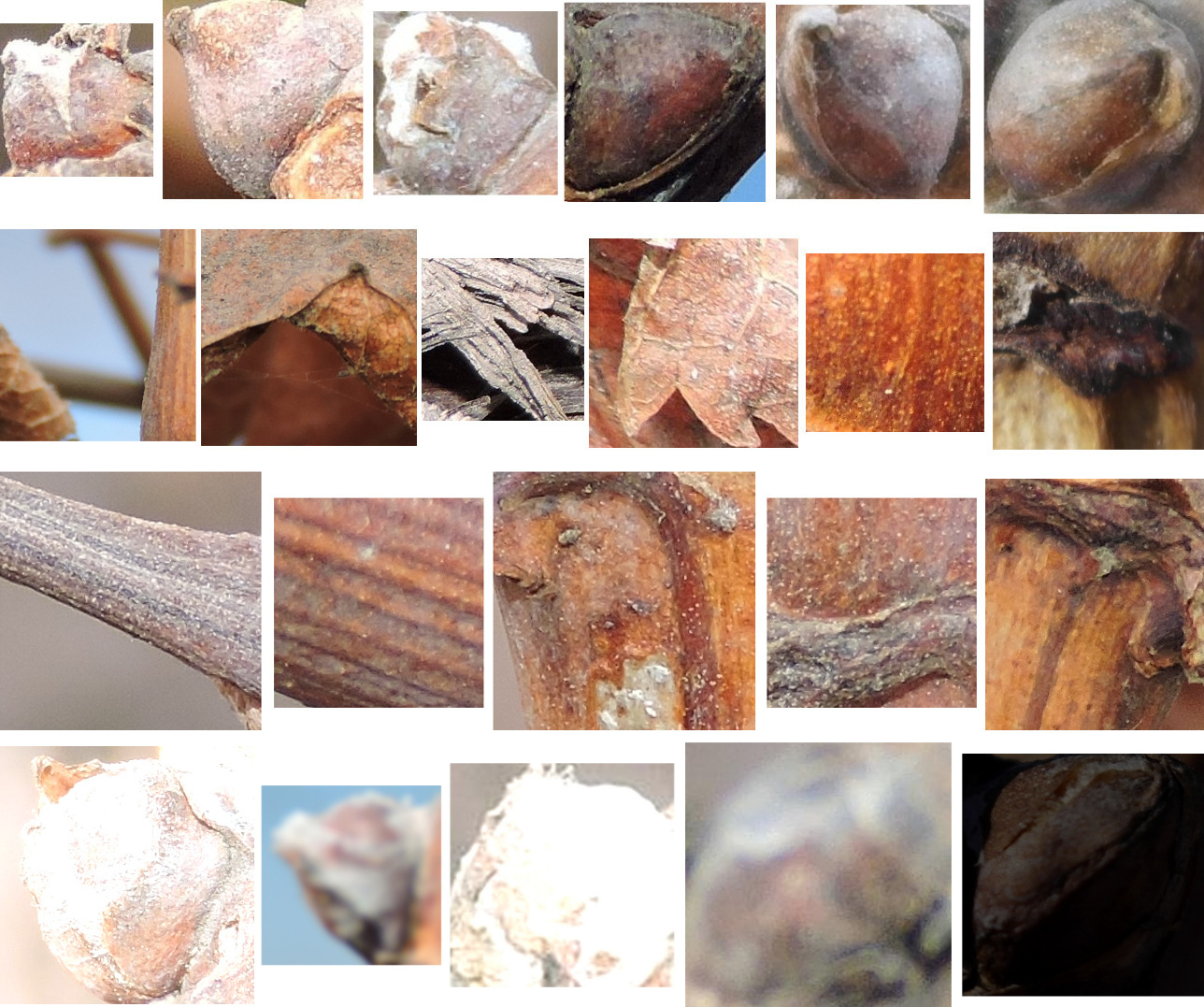}
    \caption{Corpus of patches used in this work. The first row corresponds to \emph{bud patches}. The second and third rows correspond to the \emph{non-bud patches}. The bottom row shows bud patches that were discarded because they are blurred, overexposed or underexposed.}
    \label{fig:dataset}
\end{figure}

All scenarios that we consider here were designed with the goal of encompassing all possible scenarios that a scanning-window detection algorithm might face. In all cases, classification is required over a rectangular patch. Using the original vineyard high-resolution images, we generated patches like those typically produced by scanning-window detection algorithms, where the proportion of bud and non-bud pixels varies from patch to patch depending on its size as well as its location in the original image. For training of the classifier, we consider two different categories of patches, one for each concept, called \emph{bud} and \emph{non-bud} patches respectively (see Fig.\ref{fig:dataset}). The \emph{bud patches} are rectangular regions that perfectly circumscribe the bud, resulting in patches that contain different proportions of bud and non-bud pixels depending on the deviation of the bud's geometry from a perfect circle, as illustrated by the images in the first row of Fig.\ref{fig:dataset}. In contrast, \emph{non-bud patches} consist in rectangular patches containing not a single bud pixel, corresponding to the background and other plant parts, such as sky, other plants, stems, leaves, trunks, wires, soil, and more, as illustrated by the images in the second and third rows of Fig.\ref{fig:dataset}. In our corpus, the non-bud patches were labeled as \emph{non-bud} class, while the bud patches were labeled as \emph{bud} class. The motivation for training with perfectly circumscribed buds is to get an ``ideal'' bud patch trying to maximize the proportion of bud pixels. Due to the non-rectangular shape of buds, it is inevitable to incorporate non-bud pixels when sampling a rectangular patch that contains the entire bud. The amount of non-bud pixels in the bud patches corresponds, in average over all patches in our corpus, to approximately 25\% of the pixels in the patch, and usually are grouped in areas near the patch corners. To build this corpus, patches of various geometries were extracted from the 760 vineyards high-resolution images. The shape geometry of the patches are rectangular and variable in size with resolutions ranging from $0.1$ to $2.56$ Megapixels (corresponding to patches ranging from $100\times100$ to $1600\times1600$ pixels approximately). The resulting patch corpus contains a total of 790 bud patches and more than 25200 non-bud patches\footnote{All images datasets available in \url{http://dharma.frm.utn.edu.ar/vise}}. The process for sampling both the bud and non-bud patches was performed using a free-hand selection tool, part of the image manipulation software developed by the authors\footnote{.Net Software and source code available in \url{http://dharma.frm.utn.edu.ar/vise}}. The protocol used to extract these patches was as follows:

\begin{enumerate}[(i)]
    \item \emph{Bud patches}: each bud was perfectly segmented (i.e., only bud pixels were selected) manually by the free-hand selection tool with a dual purpose, i.e. to fit the bud entirely in a rectangular patch and to create a mask of the bud, which allowed to calculate exactly the amount of bud and non-bud pixels in the patch (information used later during the evaluation of our approach). The bud patches were processed manually to discard the unfocused, overexposed or underexposed patches (see examples in the bottom row of Fig.\ref{fig:dataset}), leaving a total of 500 bud patches. We consider that discarding such cases does not violate the practicality of our approach, as in practice it is easy to avoid such image distortions either by simple visual inspection, or autonomously as exemplified by the works of \cite{hongqin2015real,lim2005detection}.
    
    \item \emph{Non-bud patches}: these have been extracted through the free-hand software tool using a semi-automated process. The user, using the free-hand tool, selects some region of the image belonging to this class. The software samples several rectangular patches, all of them within entirely the region, with a pre-selected patch step size and dimensions. This method works similarly to a scanning-window algorithm, but we limit it to scan in a restricted region. With this procedure we could obtain a lot of examples, orders of magnitude more than the bud patches. 
\end{enumerate}

\subsection{From pixels to visual descriptors}
\label{subsec:descriptors}

The next step is to compute the visual descriptors for each patch, resulting in a new corpus containing one visual descriptor per patch. This process consists of a two-steps workflow:

\begin{enumerate}[(i)]
    \item Given a patch, the first step is to calculate a set of low-level local visual features of the patch. These features were computed using the \emph{Scale-Invariant Feature Transform (SIFT)} algorithm \citep{lowe2004distinctive}, so we refer to them as \emph{SIFT features}. Below we show some insights of the SIFT algorithm, along with some implementation decisions particular to this work.
    
    \item Then, the patch visual descriptor is constructed by mapping the (variable number of) low-level features to a vector of some user-defined constant size. This vector is a high-level descriptor obtained by the \emph{Bag of Features (BoF)} algorithm \citep{csurka2004visual}, which we called \emph{BoF descriptor}. The BoF descriptor is the visual information used in the training and classification phases. Below we show some insights on the BoF algorithm, along with some implementation decisions particular to this work.
\end{enumerate}

\subsubsection{Scale-invariant feature transform (SIFT).}

\emph{SIFT} is one of the most popular algorithms for extracting visual features in images. In general, the algorithm takes an image and transforms it into a set of features vectors that are invariant to scale, rotation and translation of the image, describing local regions of the image (of tens of pixels in size). The detailed explanation of SIFT is beyond the scope of this paper, but you can see the original work of Lowe \citep{lowe2004distinctive} for all details. The main intuitions can be summarized in four steps:

\begin{enumerate}[(i)]

    \item \emph{Scale-space extrema detection}. The real-world objects only make sense in a certain scale that depends on the object size and the observer viewpoint. This \emph{multi-scale} property is very common in nature, and it is possible to approximate this concept in digital images from functions known as \emph{scale-space}. SIFT proposes a function for producing from the original image, a succession of images of different levels of blur, by first rescaling the original image into images of lower resolutions, to then convolve each pixel of each image with a Gaussian centered at the pixel. To identify potential \emph{keypoints}, \emph{difference of Gaussians (DoG)} is applied between two successive images of the same scale to obtain a \emph{DoG image}. Potential keypoints are taken as the local maxima and minima in the DoG images that occur on different scales. To detect these local maxima and minima, each pixel is compared with its 8 neighbors in the same DoG image, and their 9 neighbors of the before and after DoG image in the same scale.
        
    \item \emph{Keypoint localization}. The above step produces a lot of potential keypoints, where some are along an edge or have not enough contrast. This kind of keypoints are unstable and are not useful as features, so a refinement is needed for more useful results. To decide the discard or not of a keypoint, the algorithm computes the curvature of the DoG images in two different direction in that point. It is considered that there is an edge, and thus that the keypoint should be discarded, if the curvature is large (the one perpendicular to the edge) and the other one is small (the one along the border). If both curvatures are large enough, above a threshold, the keypoint is considered to be in a corner, and is kept. If both curvatures are small, the keypoint is located on a flat region and is also kept. Low contrast keypoints are simply discarded verifying that its intensity is below a threshold.
    
    \item \emph{Orientation assignment}. At this point, we have a set of stable keypoints invariant to scale. The next step is to assign a consistent orientation based on image local features. This step is key for achieving invariance to image rotation. For this, the algorithm takes around each keypoint a neighborhood whose size depends on the scale on which it was detected (higher scale, larger regions), and then calculates the magnitude and direction of the gradient at each pixel, to finally build an orientation histogram with 36 bins covering 360 degrees. The most prominent bin is taken as gradient of the keypoint. For any other bin exceeding 80\% of the most prominent bin, a new keypoint is created and its gradient is computed, i.e. different keypoints can be created with the same location and scale, but with distinct orientations.
    
    \item \emph{Keypoint descriptor}. At this point, there is already a set of keypoints invariant to scale with an assigned orientation. The last step is to create a descriptor or fingerprint for each keypoint that is invariant to image scale, rotation, and translation. For this, the algorithm defines a $16\times16$ pixel neighborhood around the keypoint, and divides it into 16 regions of $4\times4$ pixels. For each region, it calculates the magnitude and direction of the gradient at each pixel. Then, an orientations histogram of eight bins is created and assigned the orientation according to the largest bin of each region. To make the descriptor invariant to rotation, the algorithm subtracts the gradient directions of each region to the gradient direction of the keypoint, so that now each gradient of the region is relative to the gradient of the keypoint. To end this process, the \emph{SIFT feature vector} is generated with $16*8=128$ values forming the keypoint descriptor invariant to scale, rotations and translations.
    
\end{enumerate}

Once the SIFT vectors for different images has been calculated, we have a variable number of vectors per image, with the number of vectors depending on the particular visual characteristics of the image. However, a classification algorithm requires input information of constant size, i.e., a descriptor for the image consisting in a set of variables assignments with the same size for every image in the corpus. To get such an homogeneous size descriptor for each image given a variable number of SIFT vectors, we use the BoF algorithm as a possible alternative which we proceed to describe in the next section. 

\subsubsection{Bag of Features (BoF).}

\emph{BoF} is a very popular visual descriptor used in image classification. Basically, it produces a constant (and compact) representation of the image, when given as input a collection of local low-level features \citep{csurka2004visual}. The computation of these features is independent of BoF, and can be computed from any feature extraction algorithm available in the literature, being SIFT one of the most popular. BoF is inspired by a concept called \emph{bag of words} used originally in text documents classification. A bag of words is a vector that maintains a counter of occurrences for each word in the document, i.e. a histogram on the vocabulary used. In computer vision, a bag of visual features is a vector that maintains a counter of occurrences for low-level features given a particular vocabulary \citep{szeliski2010computer}, that must be computed from the corpus. BoF involves two main steps:

\begin{enumerate}[(i)]

    \item The first step is an offline process to build the vocabulary of visual features from a \emph{train set} of patches. To achieve a compact representation, the algorithm generates the vocabulary by clustering vectors in the visual features space through a clustering algorithm, such as \emph{k-means}. With SIFT, BoF performs the clustering according to the 128-dimensional SIFT vector for each keypoint in the train set. Then, each cluster is treated as a single visual word in the visual vocabulary. The vocabulary size is a design decision provided externally by a user.

    \item The second step of BoF consists in the histogram computation of the SIFT vectors of an image, over the vocabulary computed in the previous step. Given a set of SIFT vectors of an image, it calculates the number of vectors that falls within each cluster, considering that a vector belongs to the cluster whose center of mass is the closest, and reports this count as the size of the corresponding bin in the histogram. 
 
\end{enumerate}

While this approach is simple and does not consider geometric information of the object of interest (buds in our case), it has been shown to have excellent performance for various visual classification tasks \citep{jiang2007towards}. To learn the algorithm in depth, you can consult the paper of \cite{csurka2004visual} and Chapter 14.4 of the book \emph{Computer Vision - Algorithms and Applications} of \cite{szeliski2010computer}.
In this work we used the SIFT implementation provided in the \emph{Non-Free} module of \emph{OpenCV}\footnote{\url{http://opencv.org}} library. We left the parameters of the algorithm in their default values. If you are interested in knowing these values, you can consult the library documentation\footnote{\url{http://docs.opencv.org/3.1.0/d5/d3c/classcv_1_1xfeatures2d_1_1SIFT.html}} \citep{kaehler2016learning}, and the standard BoF implementation provided in \emph{OpenCV}. We build the vocabulary using \emph{k-means} as clustering algorithm \citep{bishop2006pattern}, and initialized the centers according to the algorithm of Arthur and Vassilvitskii \citep{arthur2007k}. There are different parameters involved in this process. The maximum number of iterations to k-means was set to $100$, and the termination criteria for desired accuracy was set to $1\times10^{-3}$. As soon as each cluster center moves less than this value in some iteration, k-means stops. During the computation of the histograms, we used the algorithm of Muja and Lowe \citep{muja2009fast} to determine efficiently the closest center to a given keypoint. All parameters of this algorithm are left in their default values. If you are interested in learning more implementation details you can consult the library documentation\footnote{\url{http://docs.opencv.org/3.1.0/d4/d72/classcv\_1\_1BOWKMeansTrainer.html}}, as well as the available literature \citep{kaehler2016learning}.

\subsection{Classification Algorithm}
\label{subsec:trainstage}

At this point, each labeled patch of the corpus has its BOF descriptor. The third stage of the approach is to train a classifier to learn the concept of bud vs. non-bud from this set of \emph{labeled} BoF descriptor, the train set. Later, the classifier is used to decide the more likely class for a set of descriptors of unclassified patches. In this work we use the \emph{Support Vector Machine (SVM)} algorithm \citep{vapnik1998statistical} to learn the binary classification model. First, we present briefly the intuitions of the SVM algorithm, and then provide some details of the training stage.

\subsubsection{Support Vector Machines (SVM).}

\emph{SVM} belongs to the family of \emph{lazy} classifiers that require the memorization of training examples, but in the case of SVM, requires only the memorization of a sparse set of examples, called the \emph{support vectors} \citep{vapnik1998statistical,bishop2006pattern}. The algorithm proceeds by constructing a hyperplane that separates optimally the decision boundaries for elements of different classes. These elements may have been projected to spaces of higher dimensionality through a \emph{kernel function}, if they are not linearly separable in their own space. Intuitively, a good separation that maximize the generalization power of the hyperplane is achieved by maximizing its margin, defined as the perpendicular distance between the decision boundaries.
In this work we used the SVM implementation provided in the \emph{LIBSVM} library \citep{chang2011libsvm}, with a \emph{Radial Basis Functions (RBF)} kernel. The parameters $\gamma$ (that specifies the kernel) and $C$ (the strength of the penalty for misclassified examples) were established at training time over a validation set, as explained in the next paragraph. The understanding of all parameters that define the SVM algorithm requires a deep theoretical understanding of the algorithm and the problem, so we invite the reader to follow the references of \cite{vapnik1998statistical} or \cite{bishop2006pattern} should he or she be interested in further details.

\subsubsection{Training of the classification model.}

For training, the SVM classification algorithm was provided with the \emph{training set} consisting in the corpus of the BoF visual descriptors of the bud and non-bud patches (c.f. \ref{subsec:datasets}), with a total of 367 bud patches and more than 25,000 non-bud patches. It is important to note that this training set is highly unbalanced in terms of the number of elements on each class, with a 1 to 68 imbalance ratio of the bud versus the non-bud class.
Imbalanced datasets are a problem for SVM, negatively affecting its generalization ability when negative examples strongly exceed in number the positive examples \citep{akbani2004applying}. A very simple approach often used to tackle the imbalance problem, consists in balancing the dataset by \emph{undersampling} the majority class, \emph{oversampling} the minority class, or a combination of both. Oversampling increases the instances number of a class by copying randomly selected elements, while undersampling reduces the instances number of a class by removing randomly selected elements (more details can be found in the work of \cite{akbani2004applying}). In this work, we balanced the train set by both oversampling the bud class, and undersampling the non-bud class, as follows: (i) we defined a \emph{balance rate} $R=\{1,2,4,8,16\}$; (ii) we defined the value $patches_{bud}=367$, i.e. the amount of bud class elements in the original train set; (iii) we performed an undersampling over non-bud class of size $R*patches_{bud}$; and (iv) we performed an oversampling over bud class of size $R*patches_{bud}$.
This results in a balanced dataset with a size larger by a factor $R$. The value of $R=1$ is the special case where only an undersampling is done for the non-bud class. Once the train set is balanced, we proceeded to train the SVM classifier, which requires the setting of the value of the $\gamma$ and $C$. Their value was selected through a $5$\emph{-fold cross-validation tuning} at training time \citep{bishop2006pattern}, over the ranges $\gamma=\{2^{-14},2^{-13},...,2^{-7}\}$ and $C=\{2^5,2^6,...,2^{14}\}$. These ranges were pre-selected by preliminary experimentation that is not discussed here. The selected values for $C$ and $\gamma$ were those that minimize the average of the validation error over the $5$-folds, where for the validation error we considered $1$ minus the \emph{f-measure} of the bud class, being the f-measure a typical assessment measure of classification problems \citep{powers2011evaluation}.

\section{Experimental Results}
\label{sec:results}

In this section we describe in detail the experimentation and the performance measures employed for assessing the quality of the proposed algorithm. We evaluate this quality by measuring the generalization error on a set of hold-on labeled patches not used during training, called hereon \emph{test set}. The test set consists of 133 bud patches and 133 non-bud patches, corresponding to elements of the original labeled corpus not used for training, i.e., they are disjoint sets, whose union produces the original corpus of labeled patches. To compute the generalization error, the classifier is run over each labeled descriptor of the test set, and its output is compared to its label. Any mismatch is counted as a generalization error in the performance measures typically used in binary classification problems \citep{powers2011evaluation}: \emph{accuracy} (Eq. \ref{eq:accuracy}), \emph{precision} (Eq. \ref{eq:precision}), \emph{recall} (Eq. \ref{eq:recall}) and \emph{f-measure} (Eq. \ref{eq:fmeasure}).
These measures are calculated based on the generalization errors carried out by the classifier: the terms \emph{true positive (tp)}, \emph{true negative (tn)}, \emph{false positive (fp)} and \emph{false negative (fn)} are counters used to contrast the class indicated by the classification of labeled descriptors with their true value class, i.e., their labels. \emph{Positive} and \emph{negative} expressions refer to the classification produced by the algorithm (with positive corresponding to buds and negatives to non-buds), while \emph{true} and \emph{false} refer to whether the classification is consistent with the actual label value or not. Intuitively, \emph{accuracy} is used to evaluate the overall effectiveness of the algorithm and corresponds to the proportion of bud and non-bud patches in the test set that has been correctly classified; \emph{recall} represents the proportion of bud patches correctly classified; \emph{precision} specifies the proportion of patches classified as bud that are really buds; and \emph{f-measure} is the harmonic mean of precision and recall. In our experiments we take the f-measure as the main analysis measure, but in cases where it is not conclusive we include in the analysis the other measures. Formally, these measures are calculated as follows:

\begin{equation}
    accuracy = \frac{tp+tn}{tp+tn+fp+fn}
    \label{eq:accuracy}
\end{equation}
\begin{equation}
    precision = \frac{tp}{tp+fp}
    \label{eq:precision}
\end{equation}
\begin{equation}
    recall = \frac{tp}{tp+fn}
    \label{eq:recall}
\end{equation}
\begin{equation}
    f \mbox{-} measure = 2 \times \frac{precision \times recall}{precision+recall}
    \label{eq:fmeasure}
\end{equation}

As explained in section \ref{subsec:descriptors}, BoF requires the size of the visual vocabulary. This value, denoted by $S$, is set by the user as there is no principled way to choose this value. In our first experiment, we show results for the classifier performance for different vocabulary sizes, with $S=\{12,25,50,100,200,350,600\}$, each averaged over ten balanced train sets of $R=1$. The undersampling process for balancing the train set discards a large number of non-bud elements (e.g., $67$ out of $68$ for $R=1$). This may strongly influence the quality of the learned classifier, benefiting or harming it by the fortuitous choice of more or less representatives elements. To mitigate the impact of the undersampling, we trained a set of classifiers for train sets of the same size (i.e., for the same $R$), each having exactly the same elements of the bud class, but different, randomly selected samples of the non-bud class. In the following evaluation we report the mean and standard deviation of the performance measures obtained by the set of classifiers trained over these different train sets.

\begin{figure}
    \centering
    \includegraphics[width=12cm]{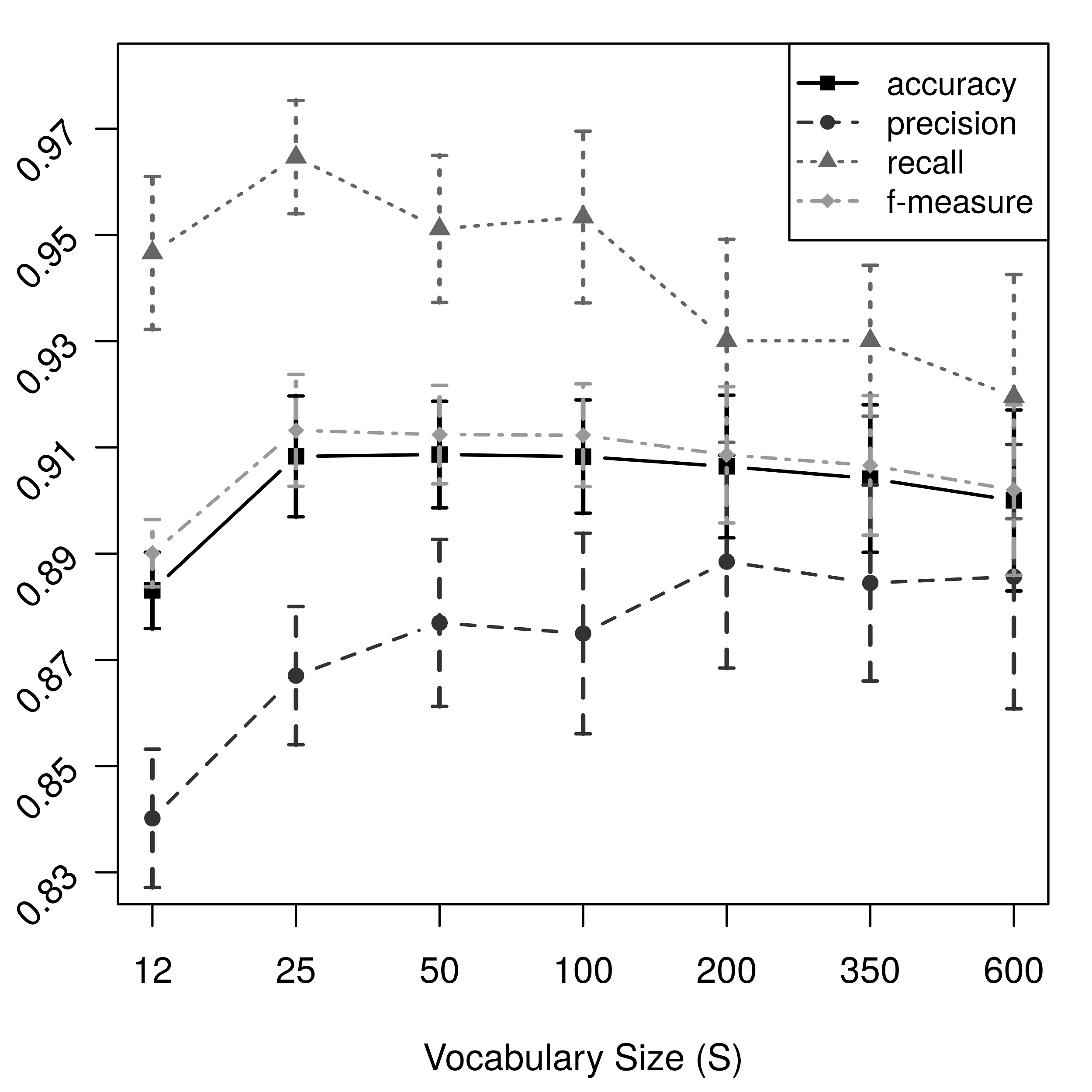}
    \caption{Accuracy, precision, recall and f-measure (mean and SD) for vocabulary size $S=\{12,25,50,100,200,350,600\}$. Vertical bars represent the SD values. All values are obtained over 10 classifiers trained for different train sets balanced with $R=1$.}
    \label{fig:measures-lineplot}
\end{figure}

The results for all four performance measures are presented in Fig.\ref{fig:measures-lineplot} and Table \ref{table:measures}, that summarize the mean and standard deviation (SD) over the $10$ trained classifiers for each $S$, all over the same test set. In the figure, the value of SD is represented by the vertical bars, while in the table it has been emphasized in italics and parentheses. For the vocabulary with $S=25$, our approach achieved a mean and \textit{SD} for accuracy of 0.908 \textit{(0.011)}, with an f-measure of 0.913 \textit{(0.011)}, recall of 0.965 \textit{(0.011)} and precision of 0.867 \textit{(0.013)}. Analyzing all vocabulary sizes, we can see that $S=25$ achieved the second worst precision while obtained the best recall and f-measure values. $S=12$ scored the worst values of accuracy, f-measure and precision. For the cases $S=\{200,350,600\}$, very similar results were achieved, but compared with $S=25$, only precision achieved slightly higher results. Finally, for the cases $S=\{50,100\}$, their results differ slightly in comparison with $S=25$: while the firsts obtain a superior precision of approximately $0.01$, their recall is below in approximately $0.012$. However, in terms of statistical significance, there is insufficient evidence to ensure that a vocabulary size is better than another, as the mean$\pm$SD of the f-measure for different values of $S$ are superposed, as can be seen in the Fig.\ref{fig:measures-lineplot}.

\begin{table}
    \centering
    \begin{tabular}{c c c c c}
        \hline
        $S$ & Accuracy & Precision & Recall & F-measure \\ \hline
        12 & 0.883 \textit{(0.007)} & 0.84 \textit{(0.013)} & 0.947 \textit{(0.014)} & 0.89 \textit{(0.006)} \\
        25 & 0.908 \textit{(0.011)} & 0.867 \textit{(0.013)} & \textbf{0.965} \textit{(0.011)} & \textbf{0.913} \textit{(0.011)} \\
        50 & \textbf{0.909} \textit{(0.01)} & 0.877 \textit{(0.016)} & 0.951 \textit{(0.014)} & 0.912 \textit{(0.009)} \\
        100 & 0.908 \textit{(0.011)} & 0.875 \textit{(0.019)} & 0.953 \textit{(0.016)} & 0.912 \textit{(0.01)} \\
        200 & 0.906 \textit{(0.013)} & \textbf{0.888} \textit{(0.02)} & 0.93 \textit{(0.019)} & 0.909 \textit{(0.013)} \\
        350 & 0.904 \textit{(0.014)} & 0.884 \textit{(0.018)} & 0.93 \textit{(0.014)} & 0.907 \textit{(0.013)} \\
        600 & 0.9 \textit{(0.017)} & 0.886 \textit{(0.025)} & 0.92 \textit{(0.023)} & 0.902 \textit{(0.016)} \\
        \hline
    \end{tabular}    
    \caption{Accuracy, precision, recall and f-measure (mean and \textit{SD} in parenthesis) for vocabulary sizes $S=\{12,25,50,100,200,350,600\}$. Numbers in bold correspond to the best mean (higher value). These values are obtained over 10 classifiers trained for different train sets balanced with $R=1$.}
    \label{table:measures}
\end{table}

\begin{figure}
    \centering
    \includegraphics[width=12cm]{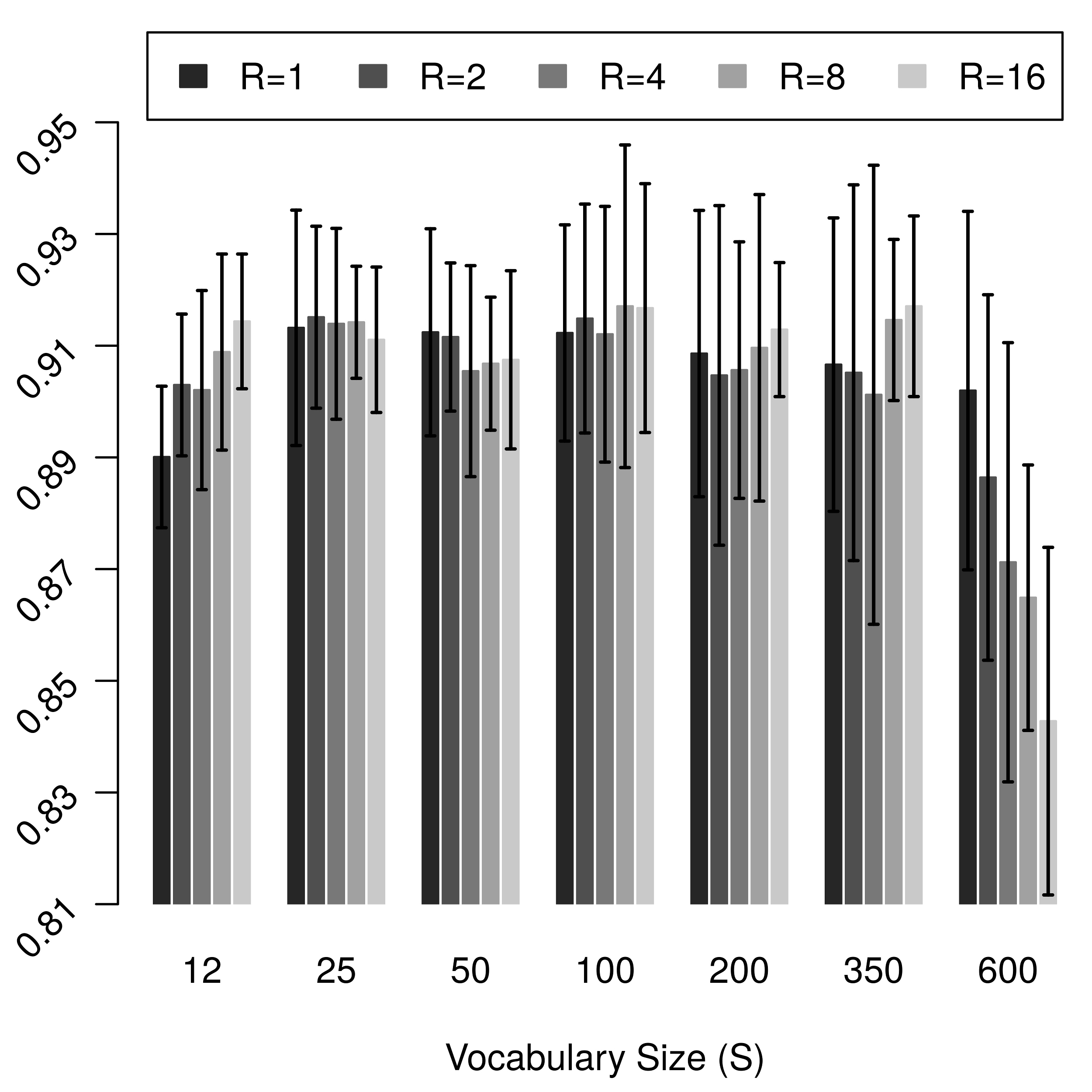}
    \caption{F-measure for different train sets obtained with $R=\{1,2,4,8,16\}$ and vocabulary sizes $S=\{12,25,50,100,200,350,600\}$. The values shown correspond to the mean and SD score of 10 classifiers trained with different train sets for each combination of $S$ and $R$.}
    \label{fig:fmTrue-mean-lines}
\end{figure}

The results presented above are calculated on the basis of different train sets balanced with $R=1$. We also evaluated the performance of the classifier for train sets balanced  over $R=\{2,4,8,16\}$. For this, we performed $10$ complete repetitions of training and testing for each combination values of $S$ and $R$, always over the same test set, reporting the mean and standard deviation of the f-measure. These results are shown in Fig.\ref{fig:fmTrue-mean-lines}, with bars representing the mean value, and standard deviation represented by the vertical interval segments. For each $S$, the figure shows a grouped of bars over all $R$, for easier comparison of the performance at each value of $R$. The figure shows that for all non-extreme values of $S$, i.e., $S \neq 12$ and $S \neq 600$, there is no  $R$ with an f-measure larger than the others, within the statistical significance bounded by its standard deviation (graphically, all intervals of mean plus/minus the standard deviation present some degree of overlap). This proves that $R > 1$ is not helpful in improving the performance of the classifier, despite the rather drastic undersampling of the no-bud class with $67$ out of $68$ elements discarded. 

\subsection{Classifier evaluation over non-bud patches}
\label{subsec:exp-categories}

\begin{figure}
    \centering
    \includegraphics[width=12cm]{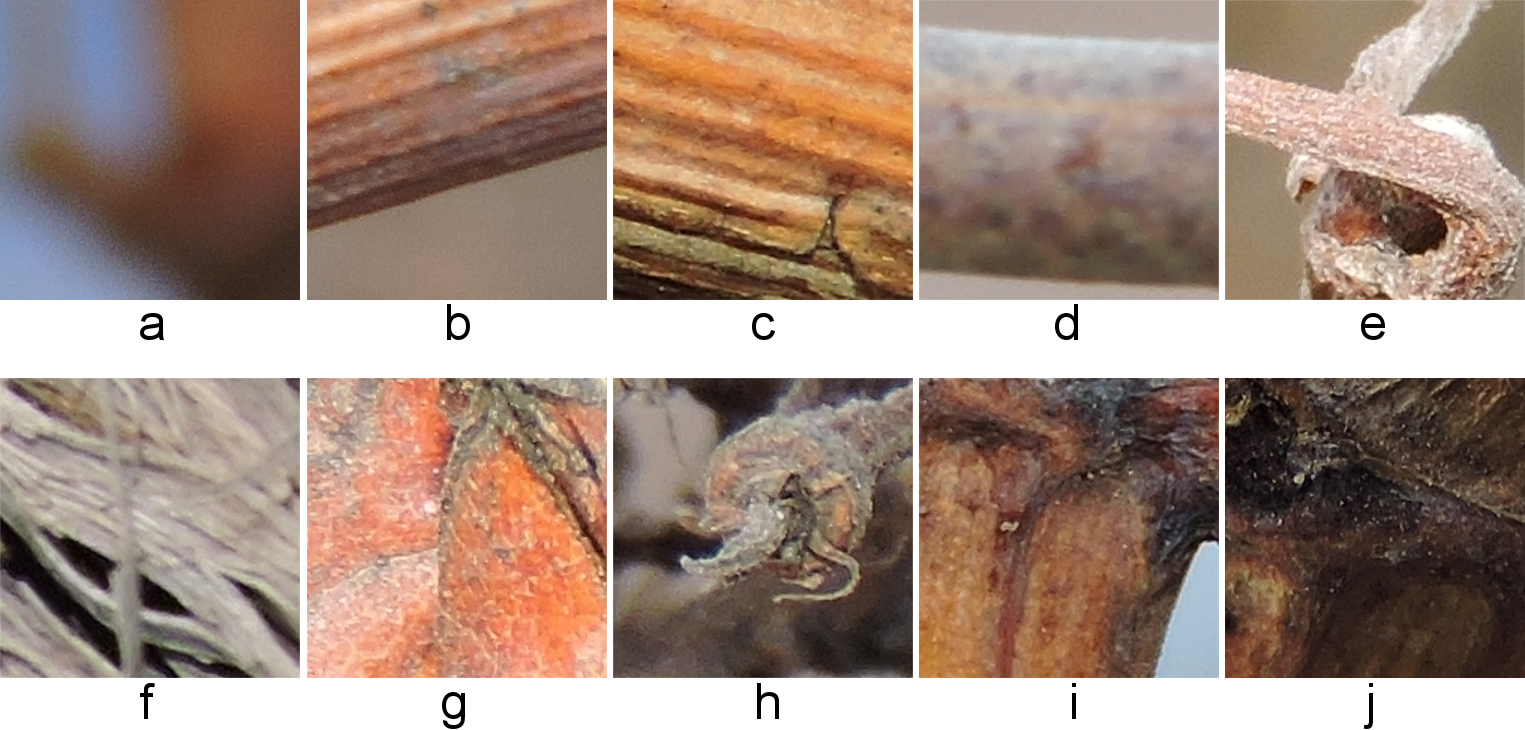}
    \caption{Actual example of non-bud patches sub-categories extracted from the testing corpus: (a) \emph{out of focus}, (b) \emph{branch edge}, (c) \emph{branch internal}, (d) \emph{wire}, (e) \emph{tendril}, (f) \emph{trunk with bark}, (g) \emph{dry leaves}, (h) \emph{dry bunches}, (i) \emph{bud neighborhood}, and (j) \emph{knot}.}
    \label{fig:category-patches}
\end{figure}

At this point, we zoom in our analysis after recognizing that the non-bud patches may be sub-categorized as parts of different, easily distinguishable regions of the background scene, that may present different levels of difficulty for the classifier in distinguishing them from the bud patches.  
We therefore designed an experiment to evaluate the classifier performance against these sub-categories. For that, we manually clustered the non-bud patches for this purpose into the following ten sub-categories (with examples shown in Fig.\ref{fig:category-patches}): (a) \emph{out of focus}: patches that are out of focus, totally blurred; (b) \emph{branch edge}: patches containing part of some branch showing an edge between the branch and the background; (c) \emph{branch internal}: patches showing only branch, arm or cane pixels; (d) \emph{wire}: patches containing some part of a metallic wire typical of field settings, (e) \emph{tendril}: patches showing total or partial tendrils, (f) \emph{trunk with bark}: patches showing partially or totally the main trunk of the plant, typically covered with the bark; (g) \emph{dry leaves}: patches showing partially or totally a dry leave typical of winter settings, (h) \emph{dry bunches}: patches showing totally or partially dry grape bunches still hanging in the plant; (i) \emph{bud neighborhood}: patches from areas close to the bud, presenting different texture irregularities and others sub-categories or combinations of them; and (j) \emph{knot}: corresponds to non-bud patches with a wood knot or part of it. 

To assess the quality of the classifier for this new test set, we used the $10$ classifiers trained during the previous experiments for each values of $S$ with $R=1$, whose training corpus contains these ten categories with their natural representativity in natural environments. To quantify the errors for each sub-category, we reported the proportion of patches in each category that were correctly classified, that is, the number of $tn / (tn+fp)$ (as the $fp$ correspond to those wrongly classified as buds), formally corresponding to the \emph{recall} measure for the non-bud class. Table \ref{table:precFalse-category} shows these quantities (recall of the non-bud class) for all ten categories, and for different vocabulary sizes, sorted in decreasing order of their mean recalls for the $S=350$ column, the one showing the best results (mostly). The most difficult categories for any vocabulary is \emph{knots}, with a non-bud recall between 0.58 and 0.68, i.e. between $42\%$ to $32\%$ of the knot patches were incorrectly classified as buds, then \emph{bud neighborhood} with non-bud recall between 0.57 and 0.73, i.e. between $43\%$ to $27\%$ of the bud neighbor patches were incorrectly classified as buds; followed by \emph{dry bunches}, \emph{dry leaves} and \emph{trunk with bark} all showing non-bud recalls between 0.798 and 0.909 for $S=350$ (considered the best case), and non-bud recalls between 0.674 and 0.888 for $S=12$ (considered the worst case). All other categories show non-bud recall values above 0.95 for most values of $S$.

\begin{table}
    \centering
    \begin{tabular}{l c c c c c c c}
        \hline
        Category         & 12    & 25    & 50    & 100   & 200   & 350   & 600 \\
        \hline
        Out of focus	 & 0.999 & 1     & 0.999 & 1     & 1     & 1     & 0.917 \\
        Branch edge      & 0.96  & 0.973 & 0.981 & 0.977 & 0.972 & 0.982 & 0.967 \\
        Branch internal  & 0.945 & 0.935 & 0.948 & 0.956 & 0.963 & 0.965 & 0.953 \\
        Wire           	 & 0.917 & 0.945 & 0.935 & 0.935 & 0.965 & 0.961 & 0.941 \\
        Tendrils         & 0.954 & 0.967 & 0.964 & 0.957 & 0.961 & 0.96  & 0.954 \\
        Trunk with bark  & 0.888 & 0.892 & 0.91  & 0.904 & 0.895 & 0.908 & 0.887 \\
        Dry leaves       & 0.793 & 0.793 & 0.806 & 0.804 & 0.834 & 0.829 & 0.815 \\
        Dry bunches      & 0.674 & 0.716 & 0.779 & 0.786 & 0.801 & 0.798 & 0.759 \\
        Bud neighborhood & 0.634 & 0.582 & 0.655 & 0.695 & 0.734 & 0.713 & 0.723 \\
        Knot             & 0.57  & 0.61  & 0.659 & 0.685 & 0.683 & 0.668 & 0.661 \\
        \hline
    \end{tabular}    
    \caption{Proportion of non-bud correctly classified (a.k.a., non-bud recall) for each of the ten sub-categories of non-bud patches, for different vocabulary sizes. The values shown correspond to the mean obtained over $10$ classifiers trained for different train sets balanced with $R=1$.}
    \label{table:precFalse-category}
\end{table}

\subsection{Classifier evaluation for scanning-window patches}
\label{subsec:exp-realistic}

\begin{figure}
    \centering
    \includegraphics[width=12cm]{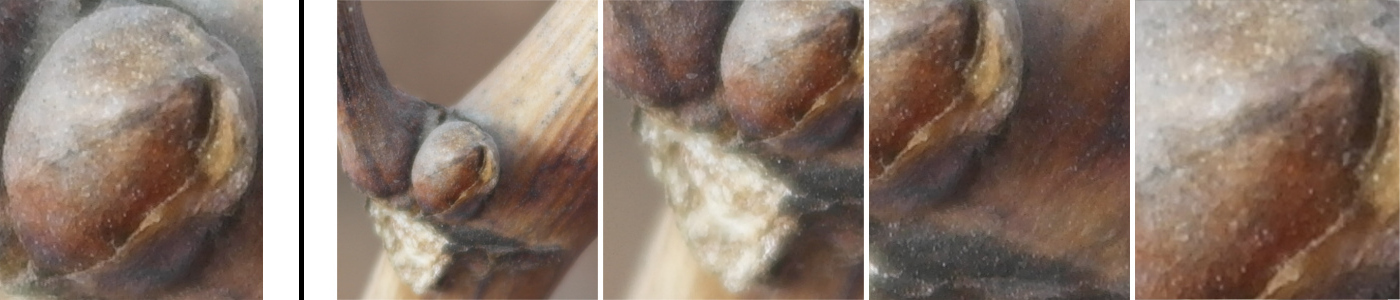}
    \caption{Left: an original \emph{bud patch} perfectly circumscribed. Right: four \emph{realistic-bud patches} obtained from the bud patch in the left. The first of these clearly has $100\%$ for the bud-pixels-kept parameters, and a small value for bud-pixels-relative. The second presents a lower value for bud-pixels-kept as some are not shown, and a larger value for bud-pixels-relative as it is closer to the bud. The third goes in the same direction, with even lower value for bud-pixels-kept and larger value for bud-pixels-relative. The last one is a case with low value for bud-pixels-kept but a value of $100\%$ for bud-pixels-relative, as only bud pixels exists in the patch.}
    \label{fig:real-patches}
\end{figure}

In the previous experiments, our classifier was tested over a corpus containing very special, unrealistic bud patches that perfectly circumscribed the buds. In the following experiment we evaluate the classifier over a more realistic testing corpus containing patches typically produced by a scanning-window algorithm, that most probably proposes patches that are far from perfectly circumscriptions of the bud. To generate these \emph{realistic-bud} patches, we implemented a script that takes each bud patch in the original test set, and adds or removes portions of non-bud pixels from the patch borders. To achieve this, our script uses the mask and bounding box information of the original bud patch, and modifies randomly the scale and/or the offset. For each bud patch in the test set, the script produces several new realistic-bud patches with different scale and offset configurations (see Fig.\ref{fig:real-patches}). For each realistic-bud patch, the number of pixels of every class can be obtained using the mask of the original bud patch. Using this information of bud pixels kept, and of new non-bud pixels added, we grouped the new realistic patches by two quantities: the \emph{bud-pixels-kept} that measures the \emph{percentage} of bud pixels kept, and \emph{bud-pixels-relative} that measures the proportion of bud pixels in the whole patch (with large values representing of patches much larger than a bud). Some combinations of these proportions were impossible to obtain, such as 100\% of bud-pixels-kept and 100\% of bud-pixels-relative, representing patches containing \emph{only} bud pixels, given the rectangular shape of the patch and the irregular shape of the bud. Also some combinations, although not impossible, were more difficult to obtain than others with our scripts, as a result, only a few examples of them were generated by the script.
 
\begin{figure}
    \centering
    \includegraphics[width=12cm]{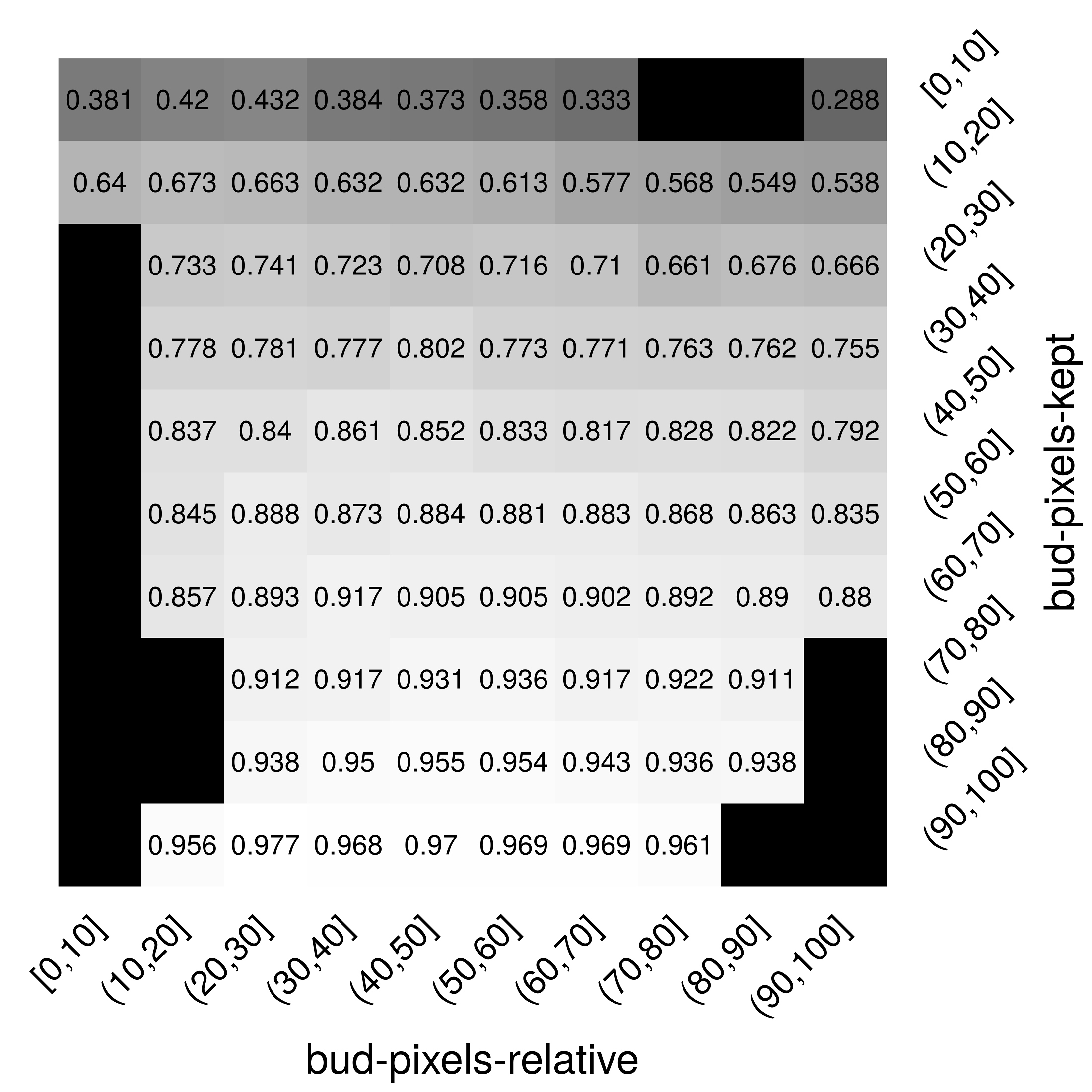}
    \caption{Heatmap of the recall in the space defined by \emph{bud-pixels-kept}, i.e. the proportion of the original bud pixels kept, and \emph{bud-pixels-relative}, i.e. the proportion of the bud pixels in the patch. This space is discretized in a regular grid of $10 \times 10$. The values showed correspond to the mean of the recall obtained over $10$ train sets of $R=1$, computed over the realistic-bud patches. Each heatmap cell contains four variations for each bud patch in the original corpus of perfectly circumscribed patches.}
    \label{fig:recall-win130k}
\end{figure}

Following we present the generalization error results achieved by the classifier for this more complex and realistic corpus of patches. The space defined by the attributes bud-pixels-kept and bud-pixels-relative was discretized in a regular grid of 10x10. For each cell of the grid, we generated $4$ realistic-bud patches from each bud patch in the original test, resulting in $133 \times 4 = 532$ realistic-bud patches per cell. Since all realistic-bud patches are labeled as bud class, only recall is evaluated, i.e. the proportion of bud patches that were correctly classified as buds, under the different patch perturbations.
To assess the quality of the classifier for this new test set, we used the 10 classifiers trained with $S=25$ and $R=1$ during the previous experiments. This choice of $S$ is justified by the fact that in the first experiments for $R=1$, the case for $S=25$ showed the best performance measures. Fig.\ref{fig:recall-win130k} shows this results in the form of a heatmap in the $10 \times 10$ discretized space, where each bin is the recall mean value achieved by the classifier for the subset of patches that fall in the range defined by the bin. In the color scale, the dark gray is the lowest value and the white is the highest, i.e., the best.
As explained, some combinations were difficult or even impossible to achieve, reaching the amount of 532 realistic-bud patches, corresponding to an homogeneous number of perturbation per original perfectly circumscribe bud-patch in only $84$ of the $100$ cells of space.
The remaining ones, i.e., those bins for which we do not generate a sufficient amount of realistic-bud patches, were discarded, and colored in black in the heatmap. As can be read from Fig.\ref{fig:recall-win130k}, the bins in the range $(20;80]\%$ of bud-pixels-relative and $(60;100]\%$ of bud-pixels-kept has a mean recall greater or equal to 0.89, which means the classifier distinguishes 9 out of 10 patches that have at least 60\% of bud pixels, and bud pixels represent only between 20\% and 80\% of the patch.

Finally, if the analysis is restricted to the dimension of bud-pixels-relative in the range $(20;80]\%$, it is observed that as bud-pixels-relative values decreases (we move toward the left columns), i.e., the size of the patch increases relative to the bud size, recall shows a tendency to increase, achieving a maximum recall of $0.977$ for patches with $(90;100]\%$ of bud-pixels-kept (lower row) and $(20;30]\%$ of bud-pixels-relative.

\section{Discussion}
\label{sec:discussion}

In this section we discuss the scope and limitations of the results obtained by the evaluation of the proposed classification algorithm, interpreting its results in the context of the scanning-window detection algorithms, including possible ways for improvement and future work.

Specifying the vocabulary size is a design decision that is set during algorithm implementation. The results presented in this paper indicate that vocabulary sizes between 25 and 100 are suitable for the classification problem of bud and non-bud patches. For $S=25$, our approach achieved a recall of $0.965$ and a precision of $0.867$, over a testing corpus containing perfectly circumscribed bud patches.
Since the perfectly circumscribed patches are atypical in scanning-window algorithms, we created a new test set of more realistic bud patches that relax the condition of perfect circumscription. In this new scenario, our classifier was robust to patches containing at least 60\% of the original bud pixels (i.e., almost half of the bud was excluded from the scanning-window), and the proportion of bud pixels in the patch was between 20\% and 80\% (i.e., between 20\% to 80\% of the pixels in the patch were non-bud pixels), achieving a recall above $0.89$ in all those cases, with an unchanged precision of $0.867$. Particularly, the best value for recall was $0.977$ for patches that hold between 90\% and 100\% of the bud pixels and these pixels represent between 20\% and 30\% of the patch, i.e. patches from three to five times larger than buds.
This behavior can be explained partially by the analysis presented in the experiment \ref{subsec:exp-categories}, about the difficulty of classification in the ten sub-categories of non-bud class. The classifier carries out the larger errors for patches of \emph{knot} and \emph{bud neighborhood} categories, in both cases patches that are in the vicinity of the bud (as knots are always located near the bud in grapevines). By increasing the patch size relative to the bud, we included visual information from non-bud categories in the bud patches, and the classifier becomes more robust. Such behavior is in line with the fact that training bud patches contains around 25\% of non-bud pixels and with works of the literature where show that including the spatial context information of the object for the classification and detection algorithms can increase the robustness of these algorithms \citep{divvala2009empirical,chen2015contextualizing,wolf2006critical,carbonetto2004statistical,heitz2008learning}. Nevertheless, using patch sizes larger than a bud in scanning-window detection algorithms can lead to displacement errors in the actual location of the bud.

On the other hand, a more detailed analysis on the characteristics of the non-bud class, shows that there are parts of the bud environment having different degrees of difficulty for the classifier. Particularly, those patches containing knots or parts close to the bud were patches where the classifier quality was negatively impacted. In the case of bud neighborhood, this behavior can be explained due to the fact that bud patches used in the training stage are rectangular regions which contain small portions of non-bud pixels (25\% in average). Then, it is expected that the low-level features computed on these regions of a bud patch, have been expressed in the patch descriptor, mistaking the classifier.
The patches with knots, presented visual features very similar to visual features that occur in the bud. One way to improve the robustness of the classifier in these situations is to use other types of low-level features, such as shape features \citep{xu2014detection}. However, the buds always develop beside knots in grapevines, so the knots can instead increase the effectiveness in bud classification, as we suggested in the above discussion on the object context information. 

There are other categories, such as \emph{dry bunches}, \emph{dry leaves} and \emph{trunk with bark}, showing non-bud recalls with mid quality results ranging from $0.67$ to $0.91$ (approximately) for the all values of $S$. In order to assess the impact of these misclassifications, we compared their representativity in the natural setting photographed in this work, with the representativity of buds, both measured as the (approximate, manually measured) proportion of patches of these categories with respect to the total number of patches in the images. We obtained representativities of $1.35\%$ for the buds; $0.35\%$ for dry bunches, $0.4\%$ for dry leaves, and $0.54\%$ for trunk with bark. We can see that all three together represent $1.29\%$, comparable that of buds, resulting in approximately $9\%$ to $33\%$ of these patches are incorrectly classified as buds, in the worst case. 
All remaining categories show non-bud recalls above $0.95$ for most values of $S$, with the case of out-of-focus deserving some further analysis. As shown in Table \ref{table:precFalse-category} for $S=25$, the non-bud recall of this category is $1$, making it clear that the classifier has no trouble in classifying them.
These patches are blurred images of different scene elements (sky, ground, other plants), and are due to the fact that we take the pictures with a focus on the buds in foreground and a small focal depth (see Fig.\ref{fig:image-bud}). In terms of the classifier, separation between ``out of focus background'' and ``focused foreground'' is somehow similar to using an homogeneously colored background, such as proposed in the work of \cite{xu2014detection} and \cite{herzog2015initial}, in the sense that the SIFT algorithm does not produce low-level features on very homogeneous or textureless regions. In our case, getting rid of focus background is not essential, but it is a situation that occurs in most of the captured images, due to the bud focusing distance and the depth of field: the closer the shooting of the bud, the lower the depth of field and the blurrier the background. In this work, with buds of at least 100 pixels in diameter, the field depth was large enough to completely blur the background. The main motivation for using images with this characteristics was not to blur the background, but to get buds on a \emph{visible scale} that can be recognized by the classification algorithm presented here, understanding that in the future, we intend to start working in a multi-scale classification and detection algorithm that allows finding buds from images where the bud is not clearly visible, i.e. when the bud occupies very few pixels in the image or even less than a pixel.

From the point of view of a scanning-window detection algorithm, detection of a particular bud is the result of the superposition of patches in a region that were classified as buds. If the classification recall decreases, the amount of bud patches misclassified increases, risking that none of the patches that include bud pixels are found, resulting in the bud not being detected. This shows that classification recall is a lower bound of the detection recall, as misclassified bud patches may still result in a detection whenever neighboring bud patches are correctly classified. For instance, the detection recall is at least $0.89$ for patches containing at least $60\%$ of the original bud pixels and the proportion of these pixels in the patch is greater than $20\%$, as shown in the corresponding cells in the heatmap of Fig.\ref{fig:recall-win130k}. This also shows that detection recall could be further improved by a scanning-window detection algorithm that has the flexibility of moving within the heatmap by varying the size and displacement step of the patch windows. By choosing the bud-pixels-kept (affected by the displacement) and bud-pixels-relative (affected by the size), a detection algorithm could easily propose a step and window size that guarantees that at some point of the scanning, the window would satisfy a bud-pixels-relative within 20\% to 30\% and a bud-pixels-kept of 90\% to 100\%, thus detecting the bud with a lower bound of $0.977$ recall (as shown by the classification recall of the corresponding cell in the heatmap). In practice, this may not be of much use, since a bud-pixels-relative within 20\% to 30\% indicate a rather unprecise localization of the bud. Nonetheless, this could be use as a good starting point for the detection algorithm, that could then move to the right of the heatmap by reducing the window size (always greater than a bud), as long as the scan guarantees that the whole bud falls within some of the windows. This procedure, although rather hypothetic, shows that the detection recall lower bound falls within the range of $0.89$ to $0.977$.

Similar to recall, a low precision means that some non-bud patches are classified as buds, resulting in detection of buds where there are none.
Our classifier achieved a precision of $0.86$, that is, $14\%$ of the non-bud patches in the corpus resulted in false positives, i.e., were wrongfully classified as bud patches.
In practice, the number of non-bud patches may largely surpasses those of the bud class, so this result may be serious depending on the application.
This analysis, however, ignores the evaluation for the sub-categories of the non-bud class shown in Table \ref{table:precFalse-category}. The complement of these values is the proportion of false positives in each category, showing that despite the underrepresented categories (dry leaves, dry bunches, and trunk with bark) and those close to the bud (knot and bud-neighborhood), all other categories show a proportion of false positives lower than $8\%$ for $S=12$ (the worst case), and lower than $4\%$ for $S=350$ (the best case); values much lower than the original $14\%$. The others are subdivided into the underrepresented, with a proportion of false positives between $9\%$ to $33\%$ but all together covering a little less number of patches as the whole bud class ($1.29\%$ versus $1.35\%$, respectively); and the knot and bud-neighborhood categories that appears to present very bad results with proportion of false positives as large as $43\%$ (for knot and $S=12$), but in reality may help a detection algorithm to detect the neighborhood of the buds, as we suggested in the above discussion on the object context information. 
So in summary, one should expect a rate of false positives detection somewhere within  $4\%$ to $8\%$. One further improvement, but rather speculative, suggests the false positives in the trunk and other categories appearing far from the buds to be sparse, occurring rather at random locations. Instead, at the vicinity of a bud one would expect a large superposition of true positive patches; suggesting a patch aggregation procedure for detection that discards isolated bud patches.

Finally, our classification algorithm for bud images is robust when the training is performed with train sets balanced according the bud class. We evaluated several scenarios for train sets balanced by undersampling of non-bud class and oversampling of bud class. The results reported for different sampling rates $R>1$ did not present improvement trends. If the computing capacity for training is a problem, the choice of $R=1$ is the most convenient, because the calculation features and descriptors with SIFT and BoF, and subsequent training SVM classifier, grows proportionally with the size of the train set.
Moreover, the CPU time incurred to classify a new patch was about 200 milliseconds (on an Intel Core i7-4770 3.40GHz processor). While this amount may seem high, we did not take into account an efficient algorithm implementation, so the SIFT features were computed for each individual patch. In practice, during the run of a scanning-window detection algorithm, we could reduce the classification time pre-calculating the SIFT features for the entire image and then extract those features that belong to each patch, saving significant amounts of computing.

\section{Conclusions}

In this work we present a classification algorithm for images of grapevine buds that involves the use of \emph{Scale-Invariant Feature Transform} for calculating low-level features, \emph{Bag of Features} for building an image descriptor, and \emph{Support Vector Machines} for training a classifier. The classification algorithm requirements are intended to be used on patches typically produced by scanning-window detection algorithms, in vineyards images taken in their natural environment. We justified the importance of bud detection through their potential applications, such as autonomous pruning systems, phenotyping of grapevines, and 3D reconstruction of the grapevines structure.
Also, we showed that the state of the art presents several limitations: (i) outdoor but uses artificial background; (ii) controlled lighting in indoor; (iii) need for user interaction; (iv) specialized stereo vision equipment; or (v) detecting buds in late-stage development, which is expected to be easier to classify because of their greater visual differentiation with the environment. In contrast, the method presented in this paper results in a contribution to the application area, as our classification approach, embedded in a scanning-window detection algorithm, is expected to be robust to more realistic scenarios, i.e. in outdoor, under natural field conditions, in winter, without artificial background, and with minimal equipment requirements.

The proposed approach achieved a classification recall greater than $0.89$ in patches containing at least $60\%$ of the original bud pixels, where the proportion of bud pixels in the patch is greater than $20\%$, and the bud is at least 100 pixels in diameter. Particularly, the best value for classification recall was $0.977$ for patches that hold between $90\%$ and $100\%$ of the bud pixels and these pixels represent between $20\%$ and $30\%$ of the patch, i.e. patches from three to five times larger than buds. A scanning-window detection algorithm could easily propose a scheme for choosing the window size and displacement that guarantees that at some point of the scanning, the window would satisfy patches with bud-pixels-kept at least $60\%$ and bud-pixels-relative greater than $20\%$. Such a scheme could start with an appropriate window size and displacement, and then manipulate these parameters, as long as the scan guarantees that the whole bud falls within some of the windows. This procedure shows that $0.89$ is a lower bound, and $0.977$ is an upper bound, for the recall of a detection algorithm.

In terms of the classification precision, the classifier achieved a value of $0.86$, that is, $14\%$ of the non-bud patches in the corpus resulted in false positives, i.e., were wrongfully classified as bud patches. Another important conclusion of this work is that for detection, it is possible to reduce the rate of false positives to somewhere within the range of $4\%$ to $8\%$, because despite the rate of false positives for classification to be $14\%$, much of these errors come from patches in the vicinity of the bud (i.e. knots and bud neighborhood patches) and patches from underrepresented categories (i.e. dry leaves, dry bunches, and trunk with bark), and should not produce errors in detection.

\section*{Acknowledgments}

This work was funded by the National Technological University (UTN), the National Council of Scientific and Technical Research (CONICET) of Argentina, and the National Fund for Scientific and Technological Promotion (FONCyT) of Argentina. We thank the the National Agricultural Technology Institute (INTA) for offering their vineyards to capture the images used in this work.

\section*{References}
\bibliography{2016-BC-arXiv-PEREZ-BROMBERG-DIAZ}

\end{document}